\documentclass[final]{cvpr}

\usepackage{times}
\usepackage{epsfig}
\usepackage{graphicx}
\usepackage{amsmath}
\usepackage{amssymb}
\usepackage[pagebackref=true,breaklinks=true,colorlinks,bookmarks=false]{hyperref}

\def\cvprPaperID{08} 
\def\confYear{CVPR 2021}

\makeatletter
\newcommand{\printfnsymbol}[1]{%
  \textsuperscript{\@fnsymbol{#1}}%
}
\makeatother

\begin{document}

\title{Delta Sampling R-BERT for limited data and low-light action recognition}

\author{Sanchit Hira\thanks{indicates equal contribution}\\
{\tt\small sanchithira76@gmail.com}
\and
Ritwik Das\printfnsymbol{1}\\
{\tt\small ritwikdas54@gmail.com}
\and
Abhinav Modi\\
{\tt\small abhinav.modi888@gmail.com}
\and
Daniil Pakhomov\\
Johns Hopkins University\\
{\tt\small dpakhom1@jhu.edu}
}

\maketitle

\begin{abstract}
   We present an approach to perform supervised action recognition in the dark. In this work, we present our results on the ARID dataset\cite{xu2020arid}. Most previous works only evaluate performance on large, well illuminated datasets like Kinetics and HMDB51. We demonstrate that our work is able to achieve a very low error rate while being trained on a much smaller dataset of dark videos. We also explore a variety of training and inference strategies including domain transfer methodologies and also propose a simple but useful frame selection strategy. Our empirical results demonstrate that we beat previously published baseline models by 11\%.
\end{abstract}

\section{Introduction}

Applications for action recognition have been on the rise with the advances in the computer vision technologies. Video action recognition finds applications in diverse areas including human-computer interaction, robotics and security surveillance. One of the important contributors to these advances is the increase in number of video datasets (Kinetics\cite{DBLP:journals/corr/KayCSZHVVGBNSZ17}, HMDB51\cite{Kuehne11}, DAVIS\cite{pont20172017}, EPIC-Kitchens\cite{damen2018scaling}, AVA\cite{gu2018ava}) which were developed for tasks such as action recognition, localization and segmentation. Most of the current work\cite{ghadiyaram2019largescale,qiu2019learning,bertasius2021spacetime, bello2021revisiting,kalfaoglu2020late} in this domain is focused on well-illuminated and high contrast datasets. However, many real-world applications of computer vision involve operation under non-ideal low contrast and lighting conditions. These include low-visibility environments, smart clothing sensors, self-driving at night, ultra-thin headset cameras, in-vivo imaging, night security surveillance etc. In this work, we focus on action recognition under dark lighting conditions.

\begin{figure}
  \includegraphics[scale=0.17]{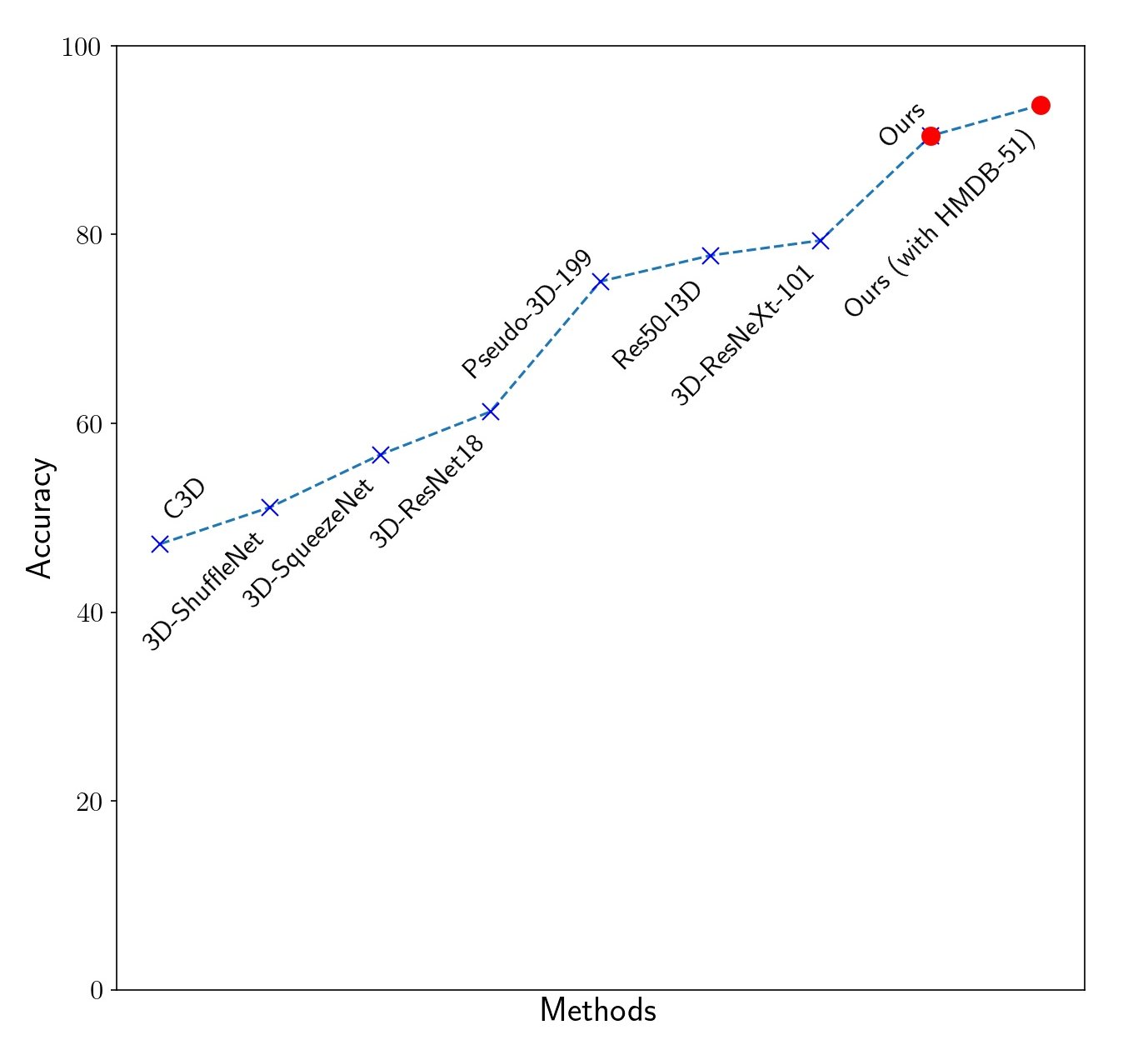}
  \caption{Performance of current CNN based approaches to action recognition vs. our approaches.}
  \label{fig:metricsoverview}
\end{figure}

In this work, we develop our strategies on the ARID\cite{xu2020arid} dataset. For a vast majority of the videos in this dataset, the illuminance is very low. This is a stark difference with other publicly available datasets like HMDB51, Kinetics-400, Charades \cite{sigurdsson2016hollywood} etc. where the action is clearly distinguishable to the human eye. To combat this we conduct experiments using multiple image enhancement methods. 

A challenging part about this dataset is the difference in distribution of the length of the clip for different action classes as well as the frame number where the action takes place. This causes the network to correlate action classes with frame positions. This leads to significantly worse performance on out of distribution test videos. Another related challenge is that for a large number of videos, the action occurs in only a small duration of the whole video. This implies that cropping clips from the larger videos and feeding them to the network might crop out the action itself. Therefore the optimal strategy would be to feed the entire video to the network. However, because of the memory constraints of the hardware and varying video lengths, this is not trivial. In this work, we present a novel sampling strategy and compare it with other methods. 

\begin{figure}
  \includegraphics[scale=0.24]{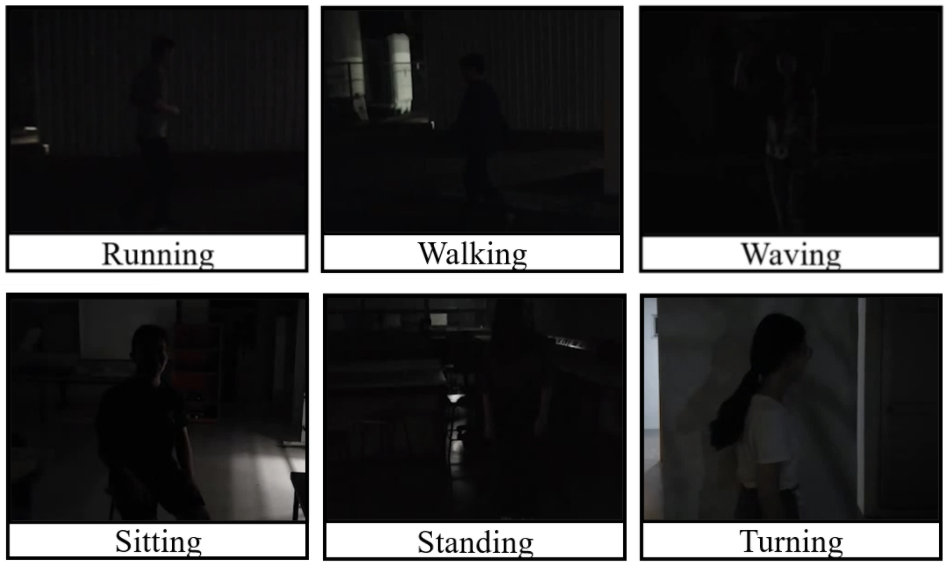}
  \caption{Sampled frames from ARID dataset, a dataset dedicated for the action recognition task in dark videos. Figure from \cite{xu2020arid}}
  \label{fig:dataset}
\end{figure}

The third challenge is that the size of the dataset under consideration is relatively small compared to other datasets. For instance, Kinetics-$400$ has $247K$ videos, EPIC-Kitchens\cite{damen2018scaling} has $28K$ training videos, Something-Something v1 \cite{inproceedings} has $86K$ videos. In comparison, this dataset only contains $1.8K$ videos. An additional compounding challenge is the low number of significantly different videos. Only $4.7\%$ of the entire dataset which is equivalent to $87$ scenes consists of unique backgrounds and human subjects. The rest $95.3\%$ of the dataset is essentially a repetition of the same subject(s) in the same background(s) doing the same corresponding action. On an average, a particular subject does the same action in the same background about 20 times with very slight variations in action performance itself. 

Solving all the above problems of dark frames, lopsided action distribution across classes and extremely small size of the dataset, in this paper we propose \textbf{Delta-Sampling R(2+1)D BERT (R-BERT)}. Our results in Figure \ref{fig:metricsoverview} demonstrate an $11\%$ improvement over previous baselines\cite{xu2020arid}.

\begin{figure*}
\includegraphics[scale=0.2]{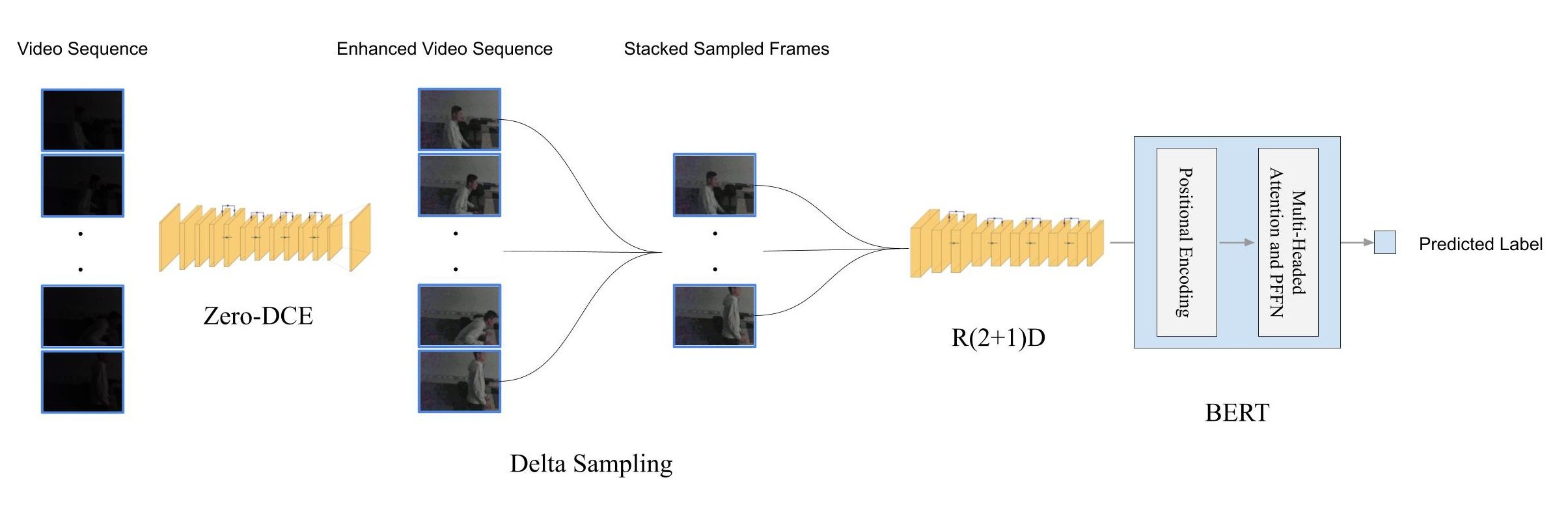}
\caption{Overview of the proposed technique. Raw frames are first extracted from provided low-light videos. These frames are then enhanced using Zero-DCE technique. The frames are then sampled using the our proposed delta sampling technique and passed to a R(2+1)D BERT network which produces the final classification.}
\label{fig:end_to_end_architecture}
\end{figure*}

\section{Related Work}
The literature analysis is divided into two different sections: (i) Advances in approaches to action recognition(\ref{action_recog_improv}) and (ii) Image Enhancement techniques(\ref{imageenhancement}).

\subsection{Advances in action recognition} \label{action_recog_improv}
Advances in the domain of action recognition can be attributed to the rise in both the number and complexity of datasets available for benchmarking. Datasets such as KTH \cite{kth} and Weizermann \cite{godrick}, which contain a relatively small number of action classes were replaced by larger and more challenging datasets like Kinetics \cite{kinetics}, UCF101\cite{ucf} and HMDB51 \cite{hmdb}. \\

A wide range of solutions have been proposed for action recognition tasks using pooling\cite{girdhar2017actionvlad, yue2015beyond}, fusion \cite{karpathy2014large} , and using Recurrent Networks to combine various temporal features. VideoLSTM\cite{li2018videolstm} performs sequential temporal modeling on 2D CNN features extracted from video frames using convolution LSTM with spatial attention. However, these methods are restricted to 2D CNNs. Also, temporal modeling was generally achieved by an additional optical flow stream or temporal pooling layers which operated on the channel dimension. 3D CNNs (C3D\cite{tran2015learning}) were also proposed for video classification and have been used for action recognition tasks. As compared to pooling and fusion techniques 3D CNNs use 3D filters to process temporal and spatial information throughout the whole network. C3D was shallow by design and thus lacked generalizability. To overcome this issue, Inception 3D (I3D) \cite{carreira2017quo} and Resnet-like versions\cite{hara2018can} of C3D were  developed. These architectures were effective but require large amounts of data for training and have huge computational and memory costs compared to their 2D counterparts. R$(2+1)$D \cite{tran2018closer} and S3D\cite{xie2018rethinking} outperformed traditional 3D CNNs by factorizing the spatial and temporal convolutions.

Other recent developments include Channel-Separated Convolutional Networks(CSN)\cite{tran2019video}, which separate channel interactions and spatio-temporal interactions in 3D CNNs. This has similarities to depth-wise separable convolutions in images. Slow-fast networks\cite{feichtenhofer2019slowfast} consisted of a slow(lower frame rate) branch and a faster(higher frame rate) branch temporally. The intuition behind this is to capture dynamics happening at different rates within the same network. The underlying architecture is a 3D CNN. 3D CNNs show promise but still lack in an effective temporal fusion strategy at the end of the architecture. This was solved by combining R($2+1$)D with Bi-directional Encoder Representations from Transformers(BERT)-based temporal modeling\cite{kalfaoglu2020late}. Recently, Bertasius \textit{et al.} \cite{bertasius2021space} have introduced a convolution-free approach to video modeling called TimeSformer\cite{bertasius2021spacetime}. It is exclusively based on space-time self-attention, suitable for long-term video modeling and achieves unprecedented results on the Kinetics-400 and Diving48\cite{li2018resound} datasets.

\subsection{Image Enhancement} \label{imageenhancement}
Conventional histogram based methods perform image enhancement by expanding the dynamic range of an image on both global\cite{coltuc2006exact} and local levels\cite{stark2000adaptive}. Many conventional methods are also proposed based on the Retinex Theory \cite{land1977retinex} which involves decomposing an image into reflectance and illumination. Fu \textit{et al.} \cite{fu2016weighted} developed a weighted variational model for estimating the illumination and reflectance.  Wang \textit{et al.}\cite{wang2013naturalness} proposed an algorithm to preserve naturalness of the image while processing images with non-uniform illumination. Guo \textit{et al.}\cite{guo2017structure} perform structure-aware smoothing by estimating a coarse illumination map and then refining it by imposing a structure prior. These methods do not take into account the noise which exists in images captured in low-light. Li \textit{et al.} proposed a Retinex model which tries to solve this issue by considering an additional noise map \cite{li2018structure}. Yuan \textit{et al.} perform global optimization to estimate an S-shaped curve for a given image for automatic exposure correction\cite{yuan2012automatic}. Most conventional methods change the distribution of image histogram or rely on potentially inaccurate physical models. This enables light enhancement on images but with the cost of creating unrealistic artifacts in the enhanced images. Advances in learning based methods were also seen in the domain of image enhancement. Bychovsky \textit{et al.} \cite{bychovsky2011dataset} introduced the first and largest dataset for supervised image tone adjustment. 

Yan \textit{et al.} \cite{yan2014learning} presented
a machine-learning based ranking approach for automatic color enhancement in an image while accounting for intermediate decisions of a human in the editing process. Yan \textit{et al.} \cite{yan2016automatic} bundled features from a pixelwise, global and contextual descriptor to achieve semantic-aware automatic photo enhancement. With advances in  deep learning, a deep autoencoder based model called ``LLNet''\cite{lore2017llnet} was proposed for low-light image enhancement.
Gharbi \textit{et al.} \cite{gharbi2017deep} introduced data dependent lookup to perform real-time image enhancement on full-resolution images. Yang \textit{et al.} \cite{yang2018image} reformulate image correction problem as an deep High Dynamic Range (HDR) mapping problem to enhance Low Dynamic Range (LDR) images. Cai \textit{et al.}\cite{cai2018learning} introduced a multi-exposure dataset and proposed a CNN based contrast enhancement method for a single image from multi-exposure images. However, most CNN based approaches are resource intensive as they require large quantities of training data. Recently, Chen \textit{et al.}\cite{chen2018deep} developed an unsupervised learning-based model using a two-way generative adversarial network (GAN) for image enhancement. Ignatov \textit{et al.}\cite{ignatov2018wespe} developed a weakly-supervised image-to-image GAN-based network, while, Deng \textit{et al.} proposed EnhanceGAN\cite{deng2018aesthetic}, an approach to image correction by adversarial learning based on binary labels on aesthetic quality. Chen \textit{et al.} \cite{chen2018learning} proposed a new dataset which includes raw sensor data to address extreme low-light image enhancement. EnlightenGAN \cite{jiang2021enlightengan} learns to enhance low-light images in an unsupervised manner by using unpaired low/normal light data. However, unsupervised GAN-based solutions usually require careful selection of unpaired training data.
Reinforcement learning was also employed to enhance
the image adjustment process \cite{hu2018exposure, park2018distort}. Recently, Guo \textit{et al.} proposed a novel method to perform Zero-Reference Deep Curve Estimation (Zero-DCE)\cite{guo2020zeroreference} thus eliminating the  need for labelled or carefully selected unlabelled data.

\section{Methodology}
\subsection{Dataset Description}
 Our dataset comprises of 6 different action classes from the ARID dataset, which are walking, running, standing, sitting, waving and turning. The task is to classify RGB videos into one of these 6 action classes. The dataset provided includes a total of 1837 videos. The training and the validation split consist of 1470 samples and 367 samples respectively. The video length in the dataset ranges from 33 to 225 frames.  

\subsection{Image Enhancement}
All of the videos in the dataset are taken at night or under very low-lighting conditions. Figure \ref{fig:dataset} gives us an insight into the dataset. To address this issue we train an enhancement method called Zero-Reference Deep Curve Estimation (Zero-DCE) for low-light image enhancement. In this method we train a model to infer dynamic range adjustments for a given image. It is equivalent to a curve estimation problem and does not require paired high and low lighting images to achieve high quality results.  In this approach, each image is a distinct training sample with no paired supervision required. A weighted, linear combination of the following loss functions is used to train the DCE Model. 

\noindent
\textbf{Spatial Consistency Loss : }
\begin{equation}
\label{equ_1}
L_{spa}=\frac{1}{K}\sum\limits_{i=1}^K\sum\limits_{j\in\Omega(i)}(|(Y_{i}-Y_{j})|-|(I_{i}-I_{j})|)^2,
\end{equation}
where $K$ is the number of local region, and $\Omega$($i$) is the four neighboring regions (top, down, left, right) centered at the region $i$. $Y$ and $I$ are the average intensity values of the local region in the enhanced version and input image, respectively.


\noindent
\textbf{Exposure Control Loss : }
\begin{equation}
\label{equ_2}
L_{exp}=\frac{1}{M}\sum\nolimits_{k=1}^M|Y_{k}-E|,
\end{equation}
where $M$ represents the number of nonoverlapping local regions of size $16\times 16$, $Y$ is the average intensity value of a local region in the enhanced image. $E$ is a hyperparameter describing the well exposedness level.

\begin{figure*}
  \includegraphics[scale=0.48]{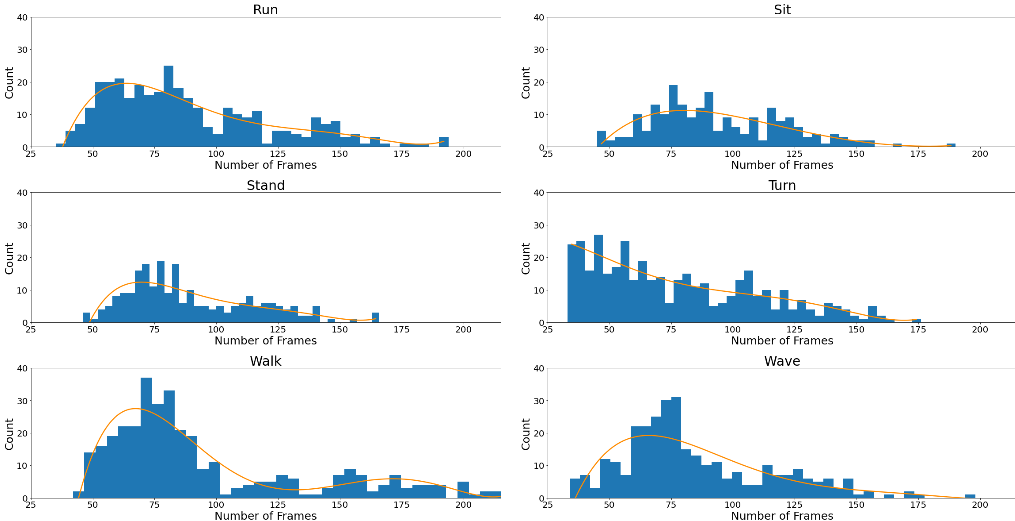}
  \caption{Distribution of frame length for each action. Differences in distribution across action classes leads to overfitting to the distribution itself instead of the action class. We solve this by proposing Delta Sampling Strategy.}
  \label{fig:frame_distribution}
\end{figure*}

\noindent
\textbf{Color Constancy Loss : }
The color constancy loss $L_{col}$ can be expressed as:
\begin{equation}
\label{equ_3}
L_{col}=\sum\nolimits_{\forall(p,q)\in\varepsilon}(J^{p}-J^{q})^2, \varepsilon=\{(R,G),(R,B),(G,B)\},
\end{equation}
where $J^{p}$ denotes the average intensity value of $p$ channel in the enhanced image and $(p, q)$ represents a pair of channels.

\noindent
\textbf{Illumination Smoothness Loss : } 
\begin{equation}
\label{equ_4}
L_{tv_\mathcal{A}}=\frac{1}{N}\sum\limits_{n=1}^N\sum\limits_{c\in\xi}(|\nabla_{x}\mathcal{A}_{n}^{c}|+\nabla_{y}\mathcal{A}_{n}^{c}|)^2,  \xi=\{R,G,B\},
\end{equation}
where $N$ is the number of iteration, $\nabla_{x}$ and $\nabla_{y}$ represent the horizontal and vertical gradient operations, respectively.


\noindent
\textbf{Total Loss.}
The total loss can be expressed as:
\begin{equation}
\label{equ_5}
L_{total}=W_{spa}L_{spa}+L_{exp}+W_{col}L_{col}+W_{tv_\mathcal{A}}L_{tv_\mathcal{A}},
\end{equation}
where $W_{col}$ and $W_{tv_\mathcal{A}}$ are the weights of the losses.

\subsection{Delta Sampling Strategy}
\label{delta_sampling_section}
The second challenge of this dataset is that the action happens only over a duration of few frames in a particular video for a large majority of the videos. A typical video consists of the following: Preparation for the action, actual action and end of action. Furthermore, the distribution of video length is different for different actions as shown in the Figure \ref{fig:frame_distribution}. This leads to overfitting, creating a bias for the frame number at which the maximum change occurs and/or to the frame distribution instead of learning the action itself. We propose a simple but useful Delta Sampling strategy to solve this problem which beats other learning-based benchmarks like SlowFast Networks and Temporal Attention as well as other sampling heuristics. 

Consider an input video $X = (X_1,..., X_N)$, $X_i$ represents the $i^{th}$ frame and $N$ represents the total number of frames in the original video. Let $T$ be the number of frames the network takes as input. We first calculate a base sampling rate for the video as $N/T$. Then we further add a $\delta$ to the calculated base sampling rate. Thus, the overall sampling rate of the video becomes $(N + T*\delta)/ T$. In our case, we sample $\delta$ to be a random floating point number from a uniform distribution in the range $[\beta, \gamma]$.

\begin{equation}
\label{equ_7}
\delta \sim Uniform [\beta, \gamma]
\end{equation}

If $\beta$ is less than zero, it is equivalent to cropping the video which leads to a much higher chance of loss of information on the important action delineation frames. Therefore in our experiments we set the lower bound of $\beta$ to be $0$. Furthermore we cap the maximum sampling rate to $\alpha$, as a very large sampling rate might cause very little information to be captured in the sampled frames.

\begin{equation}
\label{equ_6}
S_\delta = min \Big( \frac{N + T \cdot \delta}{T}, \alpha \Big)
\end{equation}
where $S_\delta$ is the adjusted sampling rate of the input video.

After the above sampling transformation, let the new video sequence be represented as $X^* = (X^*_1, ..., X^*_{N/S_\delta} )$. The input to the network is padded with blank frames on both sides. If $P_1$ and $P_2$ are the number of blank frames on both sides, then $P_1$ is sampled as follows. 

\begin{equation}
\label{equ_7}
P_1 \sim Uniform [0, T - N/S_\delta)
\end{equation}

Then, $P_2$ can be calculated as
\begin{equation}
\label{equ_7}
P_2 = T - N/S_\delta - P_1
\end{equation}

\begin{figure}
  \includegraphics[scale=0.27]{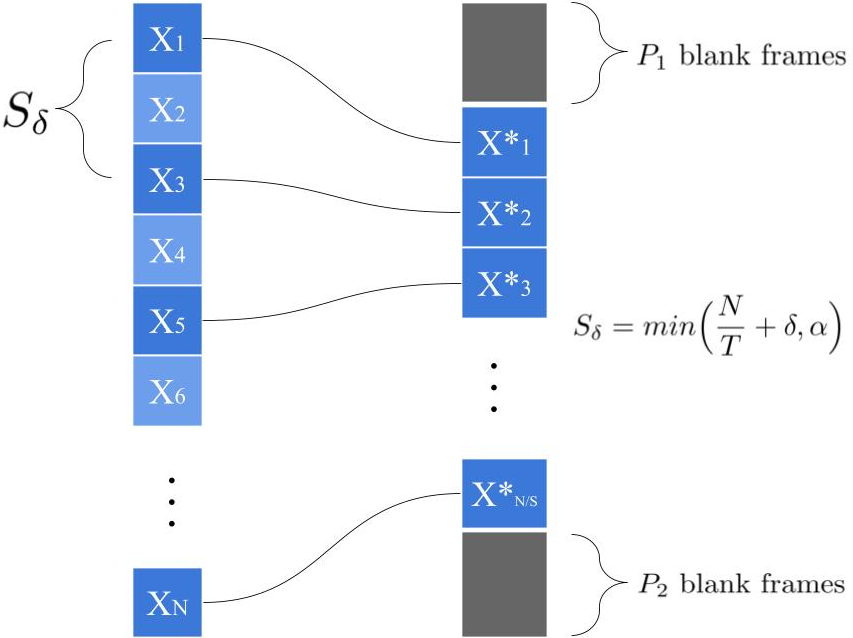}
  \caption{Frame sampling using Delta Sampling strategy. Sampling is done at rate $S_\delta$ and the sampled frames are padded with $P_1$ and $P_2$ blank frames at beginning and end respectively.}
  \label{fig:deltasampling}
\end{figure}

Therefore, we can calculate 

\begin{equation}
\label{equ_8}
\bar{X} = (P_1 \text{ }frames, X^*_1,..., X^*_{N/S_\delta}, P_2 \text{ } frames)
\end{equation}
where $\bar{X}$ denotes the final video input sequence to the network. Our approach is summarized in Figure \ref{fig:deltasampling}.

\subsection{Network Architecture}

We feed in input video sequence through image enhancement and delta sampling strategy as described in the previous sections. For the video-classification network, we use R(2+1)D - BERT  with a Resnet-34 backbone. Our findings show that the temporal attention mechanism within BERT gives a much better performance than temporal average pooling as shown in Section \ref{experiments_section}. The BERT model preserves the temporal positional information in the input features by adding a learned positional encoding. A classification embedding is also appended \cite{devlin2018bert}. After attending the features with the self-attention mechanism of BERT, a Position wise Feed Forward Network (PFFN) is applied. Finally, a simple linear layer generates the final classification output of the model.

\begin{equation}
\mathbf{y}_{\mathbf{i}}=P F F N\left(\frac{1}{N(x)} \sum_{\forall j} g\left(\mathbf{x}_{\mathbf{j}}\right) f\left(\mathbf{x}_{\mathbf{i}}, \mathbf{x}_{\mathbf{j}}\right)\right)
\label{equation:yclsbert}
\end{equation}
, where $\mathbf{x}_{\mathbf{i}}$ are the $\mathbf{i}^{th}$ temporal index of the input embeddings and $\mathbf{x}_{\mathbf{j}}$ are all possible combinations of input embeddings. $f\left(\mathbf{x}_{\mathbf{i}}, \mathbf{x}_{\mathbf{j}}\right)=\operatorname{softmax}_{j}\left(\theta\left(\mathbf{x}_{\mathbf{i}}\right)^{T} \phi\left(\mathbf{x}_{\mathbf{j}}\right)\right)$ gives the similarity between $\mathbf{x}_{\mathbf{i}}$ and $\mathbf{x}_{\mathbf{j}}$, where $\theta(\cdot)$ and $\phi(\cdot)$ are linear projections. PFFN is defined in equation \ref{equation:pffn}, where $G E L U(\cdot)$ represents the Gaussian Error Linear Unit activation function \cite{hendrycks2016bridging}.

\begin{equation}
P F F N(x)=\mathbf{W}_{2} G E L U\left(\mathbf{W}_{1} \mathbf{x}+\mathbf{b} \mathbf{1}\right)+\mathbf{b} 2
\label{equation:pffn}
\end{equation}

We train our model end-to-end with focal loss\cite{lin2017focal}. We use horizontal flipping, random rotation and random translation for video augmenting our dataset. Figure \ref{fig:end_to_end_architecture} shows the end-to-end pipeline of the proposed method.

\section{Implementation and Experiments}
\label{experiments_section}
\subsection{Image Enhancement}
\begin{figure*}
  \includegraphics[scale=0.57]{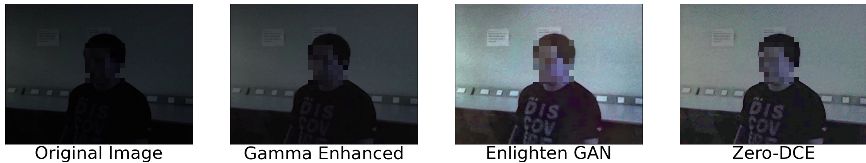}
  \caption{Visual Enhancement results (face blurred) across different methods. Zero-DCE is the most dynamically illuminated with more enhancement on the darker areas and lesser on the over-exposed areas.}
  \label{fig:enhancement_comparison}
\end{figure*}

For image enhancement, we train Zero-DCE model on $2000$ random images of different illumination and $2500$ images from the ARID dataset. We sample the $2500$ images uniformly from the videos in our training set. We use the Adam optimizer and train our model for $100$ epochs. We set $W_{tv_\mathcal{A}}$ to $200$, $W_{col}$ to 5 and $W_{spa}$ to $10$. We set the learning rate to $1e-4$. We also explore Gamma Image Correction(GIC) and EnlightenGAN to enhance the images.  However the latter two methods do not result in any improvement in the downstream action recognition task. For gamma correction we observe that a blanket gamma correction does not improve our results irrespective of the value of gamma and in-fact decreases it. We hypothesize that this is due to varying degrees of darkness across the dataset which cannot be captured properly by the gamma correction table. Upon a visual inspection, as shown in Figure \ref{fig:enhancement_comparison} we notice that EnlightenGAN had a lot of image artifacts in addition to enlightening the image uniformly with little consideration of lighting conditions at existing brighter areas of the image. Zero-DCE on the other hand solves both of these problems and the generated dynamic lighting is much more suitable to the downstream tasks. We use the trained DCE model as a frozen-network for both training and inference phases of R(2+1)-D BERT. Table \ref{table:enhancement} shows the performance benefit of using Zero-DCE as compared to other baselines.

\begin{table}
\begin{center}
\begin{tabular}{|l|l|}
\hline \textbf{Method} & \textbf{Accuracy} \\
\hline No Enhancement Baseline & 87.48 \\
\hline Gamma Image Correction & 84.62 \\
\hline Enlighten GAN & 87.50 \\
\hline \textbf{Zero-DCE Enhanced Images} & 90.46 \\
\hline
\end{tabular}
\end{center}
\caption{Comparison of different methods of enhancing brightness of low-light images.}
\label{table:enhancement}
\end{table}

\subsection{Delta Sampling Strategy}
For the Delta Sampling strategy, our network takes $T=64$ frames as input. The number of frames in the original videos vary from a minimum of $33$ to a maximum of $225$ as shown in the distribution curves in Figure \ref{fig:frame_distribution}. We set $\beta =0$, $\gamma = 1.5$ and the maximum sampling rate($\alpha) = 4$. $P_1$ and therefore $P_2$ are randomly determined at train time within their respective bounds. During test-time, we average predictions across 5 different sets of $(\delta, P_1, P_2)$ values as a test-time sampling augmentation technique. We compare our method to the following alternative approaches.
\begin{enumerate}
    \item \textbf{Slow Fast Network}: Previous works have proposed similar approaches like SlowFast network. It involves a slower and a faster network streams which operate at different temporal resolutions. Intuitonally, it tries to solve the same problem of different actions occurring at different rates across the dataset. We demonstrate that our approach is superior than the SlowFast network.
    \item \textbf{Squeeze-Excitation Module}: The second method we test our strategy against is attention along the time dimension in the backbone. For this method, we use Squeeze-Excitation \cite{hu2019squeezeandexcitation} within the Resnet-34 backbone.
    \item \textbf{Constant Sampling Rate}: In this third study, we compare our approach with a constant sampling rate of $S_{constant} = N_{max}/T$. Here $N_{max}$ is the maximum number of frames across all videos in the training set. Since $T*S_{constant} > N_{i}$ for all videos where $N_{i} < N_{max}$ we pad the remaining frames on both sides with $P_1$ and $P_2$ blank frames using the same strategy as described in Section \ref{delta_sampling_section}.
    \item \textbf{Length Adjusted Sampling Rate}: The fourth baseline we compare against is a sampling method at a video length adjusted sampling rate of $S_{len\_adj} = N_i/T$ i.e., which covers the entire $T$ frames where $N_i$ is the number of frames in the $i_{th}$ training sample. In this approach, there is no need to pad any sampled training videos since we intentionally choose $S_{len\_adj}$ to cover all $T$ frames.  
    \item \textbf{Variable Sampling Rate}: For the last sampling rate comparative study, we use a randomly chosen sampling rate $S_{var} \sim Uniform [N_{min}/T, N_{max}/T]$.  Here $N_{min}$ is the minimum number of frames across all videos in the training set. Similar to the method proposed in section \ref{delta_sampling_section} we pad the remaining frames on both sides with $P_1$ and $P_2$ blank frames whenever $T*S{var} > N_{i}$. 
\end{enumerate}
According to table \ref{table:framesampling}, the experiments show that our strategy is superior to all of the above methods.

\begin{table}
\begin{center}
\begin{tabular}{|l|l|}
\hline \textbf{Method} & \textbf{Accuracy} \\
\hline Slow-Fast Network & 44.40 \\
\hline Squeeze Excitation Module & 70.78 \\
\hline Constant Sampling Rate  & 80.93 \\
\hline Length Adjusted Sampling Rate & 85.29 \\
\hline Variable Sampling Rate & 88.28 \\
\hline \textbf{Delta Sampling Strategy (Ours)} & 90.46 \\
\hline
\end{tabular}
\end{center}
\caption{Comparison of different methods of sampling frames in videos.}
\label{table:framesampling}
\end{table}

\subsection{Network Architecture Details}
The models are trained using Ranger optimizer \cite{lessw2020ranger}, which combines Rectified Adam\cite{liu2019variance}, LookAhead optimizer \cite{zhang2019lookahead} and Gradient Centralization \cite{yong2020gradient} into one optimizer. Given the small size of the dataset we consider here, we use a small batch size of 2 which produces a regularization effect as discussed by Smith \textit{et al.}\cite{smith2018disciplined}. We train our models for 300 epochs using One Cycle Learning strategy \cite{smith2017cyclical} and focal loss \cite{lin2017focal}, as defined in the equation \ref{focal_loss}, as our classification loss. 

\begin{equation}
\label{focal_loss}
F L\left(p_{t}\right)=-\alpha_{t}\left(1-p_{t}\right)^{\gamma} \log \left(p_{t}\right)
\end{equation}

To limit the computational requirements, we use resized image frames of size $112 \times 112$ and sample $64$ frames from each video, using our delta sampling strategy discussed in section \ref{delta_sampling_section}. The output of the spatial R(2+1)D branch is a $512\times 8\times 7\times 7$ feature map, which is then average pooled in the spatial domain to reduce it to a dimension of $512\times 8\times 1\times 1$. The features are then normalized in the time domain before  passing them on to the BERT model which generates the final classification output. The BERT model uses a hidden size of $512$ with a single layer with $8$ attention heads. We use PyTorch framework \cite{NEURIPS2019_9015} for our experiments.

For training the R(2+1)D BERT model, we train on $6$ classes of low-light videos of the ARID Dataset. 
We present the following comparative studies against our strategy : 
\begin{enumerate}
    \item \textbf{3D Convolutional Networks} - We compare against C3D, 3D ResNeXt-101, 3D SqueezeNet and other 3D action recognition networks in Figure \ref{fig:metricsoverview}.
    \item \textbf{R(2+1)D Temporal Global Average Pooling(TGAP)} We take R(2+1)D network and use a temporal pooling layer at the end. We see that this leads to very interesting performance pitfalls. The network errors frequently include confusion between sitting and standing (which is sitting but in reverse), and also walking and running at times. We attribute this confusion to the loss of temporal relationship due to a global average pooling layer. Hence, we use BERT in our successive experiments as it has temporal positional encodings. 
    \item \textbf{Efficient-Net BERT} - Here, we use a modified Efficient-Net \cite{tan2020efficientnet} as the backbone. However, we observe that this network is extremely heavy and hard to train given limited computation resources. Therefore, we revert back to the lighter R(2+1)D backbone in the next two sections. 
    \item \textbf{R(2+1)D BERT with Optical Flow} - In this study, we first compute dense optical flow on the frames \cite{farneback2003two} and then pass the $2$ flow frames along with the $3$ RGB channels. We notice empirically that this experiment leads the network to hugely overfit on the training set and therefore generalize poorly. We hypothesize that this is partially due to large noise in the videos which leads to noisy optical flow values, exacerbated by the small training set size (87 unique scenes) and increase in the number of partially redundant optical flow features.
\end{enumerate}
Table \ref{table:network_arch} presents the performance on TGAP and BERT architectures with different backbones and input types as discussed above. 
\begin{table}
\begin{center}
\begin{tabular}{|l|l|}
\hline \textbf{Method} & \textbf{Accuracy} \\
\hline R(2+1)D TGAP & 82.74 \\
\hline Efficient-Net BERT  & 64.12 \\
\hline R(2+1)D-BERT with Optical Flow & 67.89 \\
\hline\textbf{Delta Sampling Resnet-BERT} & 90.46 \\
\hline
\end{tabular}
\end{center}
\caption{Comparison of different network architectures on action recognition accuracy.}
\label{table:network_arch}
\end{table}

\subsection{Additional Data}
We also present our improved results when we add corresponding data from HMDB-51. We choose to add the  videos from HMDB-51 belonging to the same 6 classes in ARID and train them in a fully supervised way. This leads to an addition of 1420 videos to our existing training set. Since the domains of both HMDB-51 and ARID are different we compare this with Multi-Representation Adaptation Network(MRAN) \cite{ZHU2019214} for cross-domain adaptation. We train MRAN with unpaired labeled videos from both source and target domains. The features extracted from the R(2+1)D model are passed to the MRAN module. The MRAN module minimizes the Conditional Maximum Mean Discrepancy (CMMD) between domains, which effectively brings the conditional distributions of the source and target domain features close to each other. The output classification of the MRAN module is then appended to the classification of the BERT module, and then final classification is obtained using a linear transform. The whole combined model is trained end-to-end using the training strategy described above. However, we observe better performance in the fully supervised experiment as shown in Table \ref{table:extra_data}.

\begin{table}
\begin{center}
\begin{tabular}{|l|l|}
\hline \textbf{Method} & \textbf{Accuracy} \\
\hline HMDB-51 + ARID - Fully Supervised & 93.73 \\
\hline HMDB-51 + ARID - MRAN  & 90.37 \\
\hline
\end{tabular}
\end{center}
\caption{Comparison of Multi-Representation adaptation network and fully supervised training.}
\label{table:extra_data}
\end{table}

\section{Conclusion}
In this work we propose Delta-sampling Resnet BERT. We show that image enhancement techniques like Zero-DCE can significantly improve the performance of machine learning models on low-light images. We also demonstrate the effectiveness of our novel efficient frame sampling strategy on skewed frame distribution across classes and low dataset volume. On the ARID dataset, we show that these techniques in combination with recent innovations in video classification architectures, improvements in deep learning optimizers and learning rate schedules can yield state-of-the-art results. We also show that simply augmenting the enhanced videos of this dataset with videos from the HMDB-51 dataset can achieve even better performance, further demonstrating generalizability of the image enhancement and sampling strategy used.



{\small
\bibliographystyle{ieee_fullname}
\bibliography{egbib}
}

\end{document}